%% file: main.tex
\RequirePackage{fix-cm}
\pdfoutput=1
\documentclass{svjour3} 

\usepackage[english]{babel}
\usepackage[utf8]{inputenc}
\usepackage{amsmath, mathrsfs}
\usepackage{dsfont, amsfonts, amssymb}
\usepackage{relsize}
\usepackage{url}
\usepackage{yaacro}
\usepackage{algorithm, algorithmic}
\usepackage[autolanguage]{numprint}
\let\n=\numprint
\usepackage{spreadtab}
\usepackage{multirow}
\usepackage{cite}
\usepackage[pdftex]{graphicx}
\usepackage{{fix-cm}}

\usepackage{hyperref}
\usepackage{xcolor}
\usepackage{algorithmic}
\usepackage{textcomp}
\usepackage{xcolor}
\usepackage{subfig}
\usepackage{gensymb}
\usepackage[numbers]{natbib}
\usepackage{caption}

\usepackage{mwe}
\usepackage{amsfonts}
\usepackage{amssymb}
\usepackage{verbatim}
\usepackage{amsbsy}
\usepackage{siunitx}
\usepackage{tabularx, booktabs}
\usepackage{soul}
\usepackage{tikz}
\usetikzlibrary{calc}

\usepackage{alphalph}
\begin{document}

\title{
Trajectory Planning for Hybrid Unmanned Aerial Underwater Vehicles with Smooth Media Transition\\
{
}

\thanks{$^{1}$Pedro M. Pinheiro, Ricardo B. Grando, Vivian M. Aoki, Dayana S. Cardoso and Paulo L. J. Drews-Jr are with the NAUTEC, Centro de Ciências Computacionais, Univ. Fed. do Rio Grande -- FURG, RS, Brazil.
E-mail: {\tt\small paulodrews@furg.br}
}
\thanks{$^{2}$A. Alves Neto is with the Electronic Engineering Dep., Univ. Fed. de Minas Gerais, MG, Brazil.
E-mail: {\tt\small aaneto@cpdee.ufmg.br}
}
\thanks{$^{4}$Alexandre C. Horn is with AuRos Robotics, Brazil.
E-mail: {\tt\small alexandredecamposhorn@gmail.com}
}
\thanks{$^{3}$César B. da Silva is with Universidade de Campinas, SP, Brazil.
E-mail: {\tt\small cesar.silva2612@gmail.com}
}
}


\author{Pedro M. Pinheiro$^{1}$, Armando A. Neto$^{2}$, Ricardo B. Grando$^{1}$, César B. da Silva$^{3}$, Vivian M. Aoki$^{1}$, Dayana S. Cardoso$^{1}$, Alexandre C. Horn$^{4}$, Paulo L. J. Drews-Jr$^{1}$
}

\input{defs}

\input{acdeff}

\begin{acgroupdef}[list=acronimos]
    \acdef{HAWV}[HUAUV]{Hybrid Unmanned Aerial Underwater Vehicle}
    \acdef[variantof=HAWV]{HAWVs}[HUAUVs]{Hybrid Unmanned Aerial Underwater Vehicles}
    \acdef{UAV}{Unmanned Aerial Vehicle}
    \acdef{AUV}{Autonomous Underwater Vehicle}
    \acdef[variantof=UAV]{UAVs}{Unmanned Aerial Vehicles}
    \acdef{UUV}{Unmanned Underwater Vehicle}
    \acdef[variantof=UUV]{UUVs}{Unmanned Underwater Vehicles}
    \acdef{ROV}{Remotely Operated Vehicle}
    \acdef[variantof=ROV]{ROVs}{Remotely Operated Vehicles}
    \acdef{PRM}{Probabilistic Roadmap}
    \acdef[variantof=PRM]{PRMs}{Probabilistic Roadmaps}
    \acdef{RRT}{Rapidly-exploring Random Tree}
    \acdef[variantof=RRT]{RRTs}{Rapidly-exploring Random Trees}
    \acdef{CL-RRT}{Closed Loop Rapidly-exploring Random Tree}
    \acdef[variantof=CL-RRT]{CL-RRTs}{Closed Loop Rapidly-exploring Random Trees}
    \acdef{PD}{Proportional Derivative}
    \acdef{PID}{Proportional Integral Derivative}
    \acdef{CNPq}{Conselho Nacional de Desenvolvimento Científico e Tecnológico}
    \acdef{CAPES}{Coordenação de Aperfeiçoamento de Pessoal de Nível Superior}
    \acdef{FAPEMIG}{Fundação de Amparo à Pesquisa do Estado de Minas Gerais}
\end{acgroupdef}

\authorrunning{P. Pinheiro et al.}
\titlerunning{Trajectory Planning for HUAUV with Smooth Media Transition\\}

\maketitle

\begin{abstract}
    In the last decade, a great effort has been employed in the study of Hybrid Unmanned Aerial Underwater Vehicles, robots that can easily fly and dive into the water with different levels of mechanical adaptation. However, most of this literature is concentrated on physical design, practical issues of construction, and, more recently, low-level control strategies. Little has been done in the context of high-level intelligence, such as motion planning and interactions with the real world. Therefore, we proposed in this paper a trajectory planning approach that allows collision avoidance against unknown obstacles and smooth transitions between aerial and aquatic media.
    Our method is based on a variant of the classic Rapidly-exploring Random Tree, whose main advantages are the capability to deal with obstacles, complex nonlinear dynamics, model uncertainties, and external disturbances.
    The approach uses the dynamic model of the \hydrone, a hybrid vehicle proposed with high underwater performance, but we believe it can be easily generalized to other types of aerial/aquatic platforms. In the experimental section, we present simulated results in environments filled with obstacles, where the robot is commanded to perform different media movements, demonstrating the applicability of our strategy.
\end{abstract}


\input{01_introduction}

\input{02_related_work}

\input{03_methodology}

\input{04_experiments}

\input{05_conclusion}

\input{06_declarations}

\section*{ACKNOWLEDGMENTS}

This work was partly supported by the \ac{CAPES}, \ac{FAPEMIG}, \ac{CNPq}, RoboCup Brazil and PRH-ANP.

\bibliographystyle{IEEEtran}
\bibliography{references}

\end{document}

%% file: defs.tex
\renewcommand{\vec}[1]{\mathbf{#1}}

\newcommand{\hydrone}{\text{HyDrone}}

\newcommand{\escalar}[1]{\ensuremath{\mathit{#1}}}
\newcommand{\vetor}[1]{\ensuremath{\boldsymbol{#1}}}
\newcommand{\matriz}[1]{\ensuremath{\mathbf{#1}}}
\newcommand{\conjunto}[1]{\ensuremath{\mathcal{\uppercase{#1}}}}
\newcommand{\aframe}[1]{\ensuremath{\left\{\mathscr{#1}\right\}}}
\newcommand{\transpose}{^{\raisebox{.1ex}{$\scriptscriptstyle T$}}}
\newcommand{\inv}{^{\raisebox{.1ex}{$\scriptscriptstyle-1$}}}

\newcommand{\Reais}[1]{\ensuremath{\mathds{R}^{#1}}}
\newcommand{\Complexo}{\ensuremath{\mathds{C}}}
\newcommand{\SE}[1]{\ensuremath{\textrm{SE}(#1)}} 
\newcommand{\SO}[1]{\ensuremath{\textrm{SO}(#1)}} 

\newcommand{\worldFrame}{\ensuremath{\aframe{W}}}
\newcommand{\bodyFrame}{\ensuremath{\aframe{B}}}

\newcommand{\xv}{\ensuremath{\vetor{x}}}
\newcommand{\pv}{\ensuremath{\vetor{p}}}
\newcommand{\vv}{\ensuremath{\vetor{v}}}
\newcommand{\rv}{\ensuremath{\vetor{\psi}}}
\newcommand{\qv}{\ensuremath{\vetor{\omega}}}

\newcommand{\uv}{\ensuremath{\vetor{u}}}
\newcommand{\motorSpeed}{\ensuremath{\Omega}}
\newcommand{\Ov}{\ensuremath{\vetor{\motorSpeed}}}
\newcommand{\ithOv}[1]{\ensuremath{\Ov_{#1}}}
\newcommand{\dwT}{\ensuremath{\delta_{T}}}
\newcommand{\dwV}{\ensuremath{\delta_{V}}}
\newcommand{\dwphi}{\ensuremath{\delta_{\phi}}}
\newcommand{\dwthe}{\ensuremath{\delta_{\theta}}}
\newcommand{\dwpsi}{\ensuremath{\delta_{\psi}}}

\newcommand{\reference}[1]{\ensuremath{{#1}_{\textrm{ref}}}}

\newcommand{\erro}[1]{\ensuremath{\escalar{e}_{#1}}}
\newcommand{\controlgain}{\ensuremath{\lambda}}
\newcommand{\Kp}{\controlgain_{p}}
\newcommand{\Kd}{\controlgain_{d}}
\newcommand{\Ki}{\controlgain_{i}}
\newcommand{\Kv}{\controlgain_{v}}
\newcommand{\Ka}{\controlgain_{\alpha}}

\newcommand{\fv}{\ensuremath{\vetor{f}}}
\newcommand{\ithfv}[1]{\ensuremath{\Fv{}_{#1}}}
\newcommand{\Fv}{\ensuremath{\fv_{\density}}}
\newcommand{\mv}{\ensuremath{\vetor{\tau}}}
\newcommand{\Mv}{\ensuremath{\mv_{\density}}}

\newcommand{\mass}{\ensuremath{m}}
\newcommand{\addedMass}{\ensuremath{\function{\mass_{a}}{\density}}}
\newcommand{\totalMass}{\ensuremath{\mass_{\density}}}

\newcommand{\inertia}{\ensuremath{\matriz{J}}}
\newcommand{\addedInertia}{\ensuremath{\function{\inertia_{a}}{\density}}}
\newcommand{\totalInertia}{\ensuremath{\inertia_{\density}}}

\newcommand{\vol}{\ensuremath{\escalar{V}}}
\newcommand{\wingSpan}{\ensuremath{\escalar{l}}}

\newcommand{\dragP}{\ensuremath{\matriz{C}}}
\newcommand{\dragR}{\ensuremath{\matriz{N}}}

\newcommand{\density}{\ensuremath{\rho}}

\newcommand{\grav}{\ensuremath{\vetor{g}}}
\newcommand{\gravTerm}{\ensuremath{\vetor{g}_{\density}}}
\newcommand{\gravStability}{\ensuremath{\vetor{c}_{\density}}}
\newcommand{\distcgtocb}{\ensuremath{\escalar{c}}}

\newcommand{\motorGain}{\ensuremath{\zeta_{\density}}}
\newcommand{\motorDrag}{\ensuremath{\eta_{\density}}}

\newcommand{\Rbtow}{\ensuremath{\matriz{R}_{\rv}}}
\newcommand{\Bwtob}{\ensuremath{\matriz{B}_{\rv}}}

\newcommand{\energy}{\ensuremath{\escalar{E}}}

\newcommand{\function}[2]{\ensuremath{#1\hspace{-.06cm}\left(#2\right)}}
\newcommand{\diagonal}[1]{\ensuremath{\textrm{diag}\hspace{-.06cm}\left[#1\right]}}

\newcommand{\InertialMatrix}{\ensuremath{\IMatrix}}
\newcommand{\InertialAddMatrix}{\ensuremath{\function{\IMatrix_a}{\flowDensity}}}

\newcommand{\ITM}{\ensuremath{\widehat{\IMatrix}}}
\newcommand{\InertialTotalMatrix}{\ensuremath{\function{\ITM}{\flowDensity}}}

\newcommand{\tree}{\conjunto{T}}
\newcommand{\bestpath}[1][]{\conjunto{P}_{#1}}
\newcommand{\Xconj}{\ensuremath{\conjunto{X}}}
\newcommand{\Xfree}{\Xconj_{\textrm{free}}}

\newcommand{\Dtc}{\ensuremath{T}}
\newcommand{\rgoal}{\refv_{\textrm{goal}}}
\newcommand{\nrv}{\escalar{{n}_{r}}}
\newcommand{\refv}{\ensuremath{\vetor{r}}}
\newcommand{\sv}{\ensuremath{\refv_{\textrm{samp}}}}
\newcommand{\node}[1][]{\ensuremath{\escalar{N}_{#1}}}

\newcommand{\transLimiar}{\ensuremath{\mu}}

%% file: 01_introduction.tex
\section{Introduction}
\label{sec:introduction}

Hybrid aerial-underwater systems have been increasingly explored in the last couple of years \cite{tan2021survey}. Vehicles capable of acting both in the air and under the water offer a large number of applications in a variety of scenarios, some of them extreme, ranging from the extreme cold of polar regions to the hot and humid climate of rain forests. Operating and transiting between air and water in such challenging environments are difficult tasks, with many setbacks to achieve a good perception and actuation overall \cite{Grando21}.
Most of these \ac{HAWVs} were inspired by aerial vehicles, such as quadcopters and hexacopters \cite{Drews2014Hybrid}. These derivations benefit the motion control in the air and underwater, generally eliminating the need for changes in their mechanics and geometry characteristics. On the other hand, the problem is that a design based on aerial vehicles tends to present a considerably better performance in the air than in the water medium. The opposite is also true for those based on underwater vehicles.  

Specifically, hybrid vehicles based on the aerial quadrotor platform are good examples of how the bias for a specific medium influences the overall performance \cite{Drews2014Hybrid, Maia2017Design, mercado2019aerial, Alzubi2018LoonCopter, Lu2018}. The structure of multirotor \ac{UAVs} generally offers a relatively easy control strategy to perform the actuation and many researchers have already taken advantage of it to design their own prototypes of \ac{HAWVs}. However, as discussed in our previous work \cite{horn2020novel}, conventional motion strategies are inefficient when deploying \ac{UAV}-like platforms under the water medium. The high density of the aquatic medium generates intense drag forces and disturbances \cite{Horn2019Study}.

\begin{figure}[!htb]
    \centering
    \includegraphics[width=\linewidth]{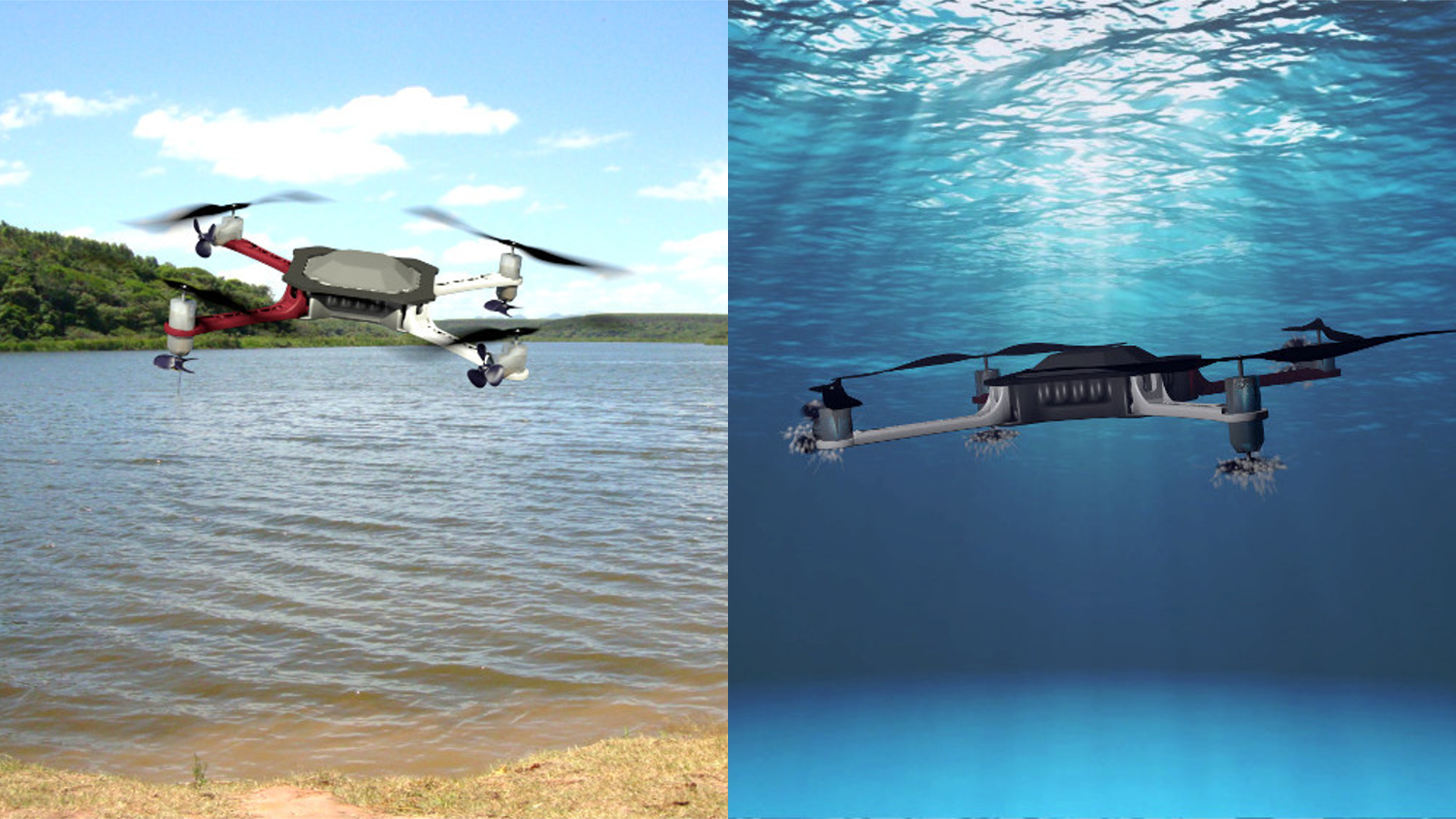}
    \caption{Simulation of the vehicle \hydrone, our proposed concept of HUAUV for aerial and underwater navigation.}
    \label{fig:proposal}
\end{figure}

In this paper, we bring the novel concept of \ac{HAWVs} that we presented in our previous work \cite{horn2020novel}. Furthermore,  we present a new trajectory planning algorithm that takes into consideration the medium transition and the vehicle's dynamics. Besides that, we discuss with more detail our previously proposed model of a hybrid vehicle (\hydrone) \cite{horn2020novel}, which has a focus in both mediums, seeking to achieve a good aerodynamic and hydrodynamic performance with the same structure. 
We show that our proposed planning algorithm manages to successfully perform the navigation through the distinct environments and avoid collision with obstacles. Figure \ref{fig:proposal} shows our vehicle navigating in the environments.

The main contributions of this work are listed as follows:
\begin{itemize}
    \item we present more details about the modelling and control of the \hydrone, an \ac{HAWV} proposed in \cite{horn2020novel} for better aquatic performance.

    \item we present a trajectory planning method for \ac{HAWVs} that takes into account the problem of media transition. The generated trajectories are smooth, in the sense that they keep the robot in hovering mode when transiting from air to water, or vice-versa.

    \item we demonstrate, with simulated experiments using the \hydrone, that our set of algorithms is successful in navigating the robot to a goal region, even when subjected to unknown static obstacles.
\end{itemize}

To the best of our knowledge, this is the first \emph{high-level} planner for \ac{HAWVs}. Other papers in the literature deal with the transition problem in terms of low-level control analysis, or simply ignored the question. Knowing that abrupt changes in the environment density may cause problems such as control instability, impact on mechanical devices, current peaks in motors, and sensor interference, we use a trajectory planner to reduce the impact of media change.

The paper has the following structure: Sec.~\ref{sec:related_work}) shows the related works; we explain the dynamic modeling of our proposed vehicle, and the set of planning algorithms in the Sec.~\ref{sec:methodology}; Sec.~\ref{sec:experiments} shows the simulated experiments; and Sec.~\ref{sec:conclusion}) shows our discussion, outlining the future works.

%% file: 02_related_work.tex
\section{Related Work}
\label{sec:related_work}

The study of \acd{HAWVs} is an emerging field in Mobile Robotics. In the last decade, the number of contributions to this field has increased significantly, with many proposed concepts and prototypes. Yet, it still has many challenges that need to be addressed, such as media transition control, underwater communication, sensor fusion, and planning strategies.

From the first works regarding \ac{HAWVs} \cite{Drews09, Drews2014Hybrid, AlvesNeto2015Attitude, Maia2015Demonstration} to recent studies \cite{horn2020novel, mercado2019aerial,Romulo18}, no major contributions were added to the high-level motion planning of quadrotor-based \ac{HAWVs} branch. In recent work, Grando \textit{et al.}\cite{Grando21} presented a navigation strategy based on reinforcement learning to \ac{HAWVs}. Nevertheless, their method is limited to pre-trained cluttered environments. Planning is one of the challenges associated with this class of vehicles, given the need to efficiently navigate and transit between two fluids that are three orders of magnitude apart with respect to specific mass.

Some studies have been conducted on modeling and control of \ac{HAWVs}, as in \cite{AlvesNeto2015Attitude} and \cite{Romulo18}, wherein the authors propose a robust attitude control with provable global stability and a comparative analysis of nonlinear state estimation methods, respectively. 
Additionally, in \cite{Maia2017Design} and \cite{mercado2019aerial}, the authors propose firstly a straightforward control scheme for a proof-of-concept prototype, and secondly a more advanced control system also adding the use of quaternions for orientation representation.
Finally, in \cite{ravell2018modeling} the authors present a hybrid controller designed for trajectory tracking considering the full system and preliminary experimental results, making this work as far as our knowledge goes, one of the few to address the smooth transition control. But trajectory planning is out of scope for all of them.

The motion planning problem is not new to Mobile Robotics, having many classical approaches and proposed solutions \cite{khatib1986real, Siegwart04Introduction, cai2009information}. 
However, an approach that considers the vehicle dynamics is a key element to obtain a smooth performance, especially during the media transition. On the other hand, environments filled with obstacles, and more specifically with different navigation densities, are hard to address.

One known approach that can handle nonlinear dynamics and complex workspaces is the \ac{RRT} \cite{LaValle01Randomized}. In this method, the authors generate random trees that rapidly explore the environment by a robot model in an open- loop. This seminal work by Lavalle and Kuffner was later the basis for many approaches like the one with a closed-loop prediction by Kuwata \textit{et al.} \cite{Kuwata2009Realtime}, called \ac{CL-RRT}, and the less resource-consuming one by Karaman and Frazzoli \cite{karaman2010incremental}. Particularly, the \ac{CL-RRT} is interesting because it allows real-time planning since it has an execution loop. It also is more robust to model uncertainties, external disturbances and even the presence of dynamic obstacles \cite{Luders10Bounds}.

\ac{RRTs} have already applied to hybrid systems, whose dynamics vary abruptly, often in the function of external events \cite{Bak2017Challenges, Wu2020R3T}. For some given conditions, they have proved to be probabilistic complete, an important topic for randomized planning methods.

Other hybrid robots' studies have applied different techniques to deal with the motion planning problem. Wu \textit{et al.} \cite{wu2019dynamics} applied an adaptive and global-best guided cuckoo search (CS) algorithm, called improved CS (ICS) algorithm to plan for a water diving transition. 
Additionally, in a work by Wu \cite{wu2019coordinated}, a coordinated path planning between a \ac{HAWV} and an \ac{AUV} was developed using a particle swarm optimization algorithm. 
Finally, Wu et al. \cite{wu2020multi} proposed an improved teaching-learning-based optimization (ITLBO) algorithm to strengthen the influence of individual historical optimal solutions in an environment-induced multi-phase trajectory optimization problem focused on the underwater target-tracking task. 
Recently, Su \textit{et al.}, \cite{Su21} improved the (ITLBO)-based trajectory optimization changing the optimization function. The previous works in the field do not consider obstacles, and they are not suitable for real-time application. Thus, we are proposing a trajectory planning capable of (re)calculate in real-time and capable to deal with obstacles in the environment.

Concerning the current literature in \ac{HAWVs}, it is possible to see that most of the papers are concentrated on low-level aspects of robotics, ranging from platform design to control stability, especially concerning the transition problem. Therefore, in the sequence, we present a trajectory planning strategy, based on the \ac{CL-RRT}, to deal with the trans-media navigation at a high level, with all advantages inherited from randomized planners. 

%% file: 03_methodology.tex
\section{Methodology}
\label{sec:methodology}

In this section, we describe our trajectory planning strategy for \acd{HAWVs}. To do this, we used the vehicle \hydrone, a hybrid vehicle that outperforms other platforms in underwater navigation while keeps the advantages of quadrotor-like aerial vehicles \cite{horn2020novel}. In the sequence, we describe the vehicle dynamics, control laws, and switching strategy used in the planner.

\begin{figure}[t]
    \centering
    \includegraphics[width=.7\linewidth]{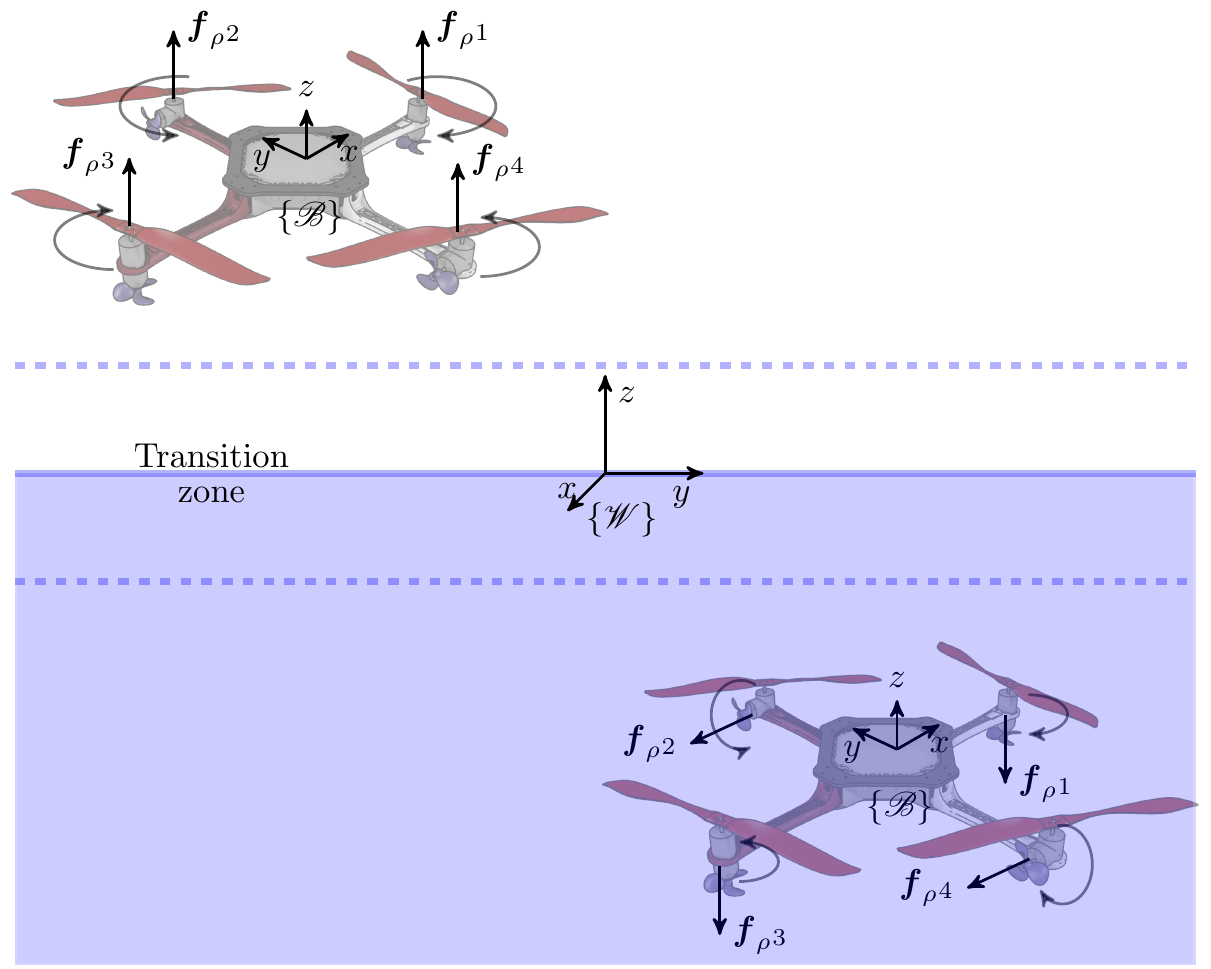}
    \caption{Forces and moments generated by the propellers of the vehicle \hydrone. Above the vehicle operates in the aerial mode, while below it acts in aquatic mode. Around of the surface level ($z = 0$) there is a transition zone \{-\transLimiar, \transLimiar\}.}
    \label{fig:model}
\end{figure}

Fig.~\ref{fig:model} presents both, aerial and underwater operation modes. In the air, aerial propellers generate forces and moments that make the robot acts like a typical quadcopter \ac{UAV}, with $\pv$ and angular $\rv$ movements in $\Reais{3}$ occurring in the function of the unbalance in the speed $\Ov$ of the rotors. On the other hand, in the water, aquatic propellers operate similarly to a \ac{ROV}, with two propellers generating forces $\Fv$ along the $x$-axis of the vehicle body, and the other two responsible for providing the thrust along $z$-axis.
Also, between both media, there is a \emph{transition zone} in which the robot only navigates up or down, as will be further explained.

\begin{problem}
    Let a \ac{HAWV} ruled by dynamics \eqref{eq:linear_speed_air} to \eqref{eq:angular_accel_air} in aerial operation mode and \eqref{eq:linear_speed_wat} to \eqref{eq:angular_accel_wat} in aquatic operation mode, with their respective control laws for each media. This vehicle navigates in a cluttered environment $\Xconj$, filled with unknown static obstacles. Let us also assume the existence of a transition zone $-\transLimiar \leq z \leq \transLimiar$, for some $\transLimiar > 0$.
    Then, the main objective is to find a trajectory \emph{in real-time} from the starting location of the vehicle to a given goal $\rgoal$, such that $\xv(t) \in \Xfree$ for all $t > 0$. The planner algorithm must also guarantee a smooth transition between air and water within the transition zone.
    \label{prob:main}
\end{problem}

\subsection{Aerial dynamics and control}
\label{subsec:aerial_dynamics}

Firstly we describe the mathematical model for the aerial operation mode. Being $\xv \in \Reais{12}$ the state vector and $\uv \in \Reais{4}$ the input vector, we have:
{\setlength{\arraycolsep}{3pt}
\begin{equation*}
    \xv =
    \begin{bmatrix}
        \pv & \rv & \vv & \qv
    \end{bmatrix}\transpose{}
    \text{~and~~~}
    \uv =
    \begin{bmatrix}
        \dwT & \dwphi & \dwthe & \dwpsi
    \end{bmatrix}\transpose{},
\end{equation*}
}
\noindent where $\vv$ and $\pv$ represent the vehicle's center mass velocity and linear position, respectively. The first one is given in meter per second and the second one in meters, where both values are relative to the world's reference frame $\worldFrame \in \Reais{3}$. Also, $\qv$ represents the angular velocity vector (in rad/s) relative to the body's reference frame $\bodyFrame$ attached to the center of mass of the \ac{HAWV}. For the last, $\rv$ is the orientation vector in the Lie Group $\SO{3}$ (in radians), relative to $\worldFrame$ as $\vv$ and $\pv$.
At the same time, inputs $\dwT$, $\dwphi$, $\dwthe$, and $\dwpsi$ are functions describing the influence of the rotor speed vector $\Ov \in \Reais{4}$ (in rpm) on the vertical thrust force and angular roll, pitch and yaw moments, respectively\footnote{Actuator dynamics are neglected since they are faster than variations on the states.}.

The final vehicle's dynamics can be written as follows:
\begin{align}
    \dot{\pv} &= \vv, 
    \label{eq:linear_speed_air}\\
    \totalMass \dot{\vv} &=
    - \Rbtow \dragP |\vv| \vv
    - \qv \! \times \! \mass \vv
    - \mass \grav
    + \Rbtow 
        \begin{bmatrix} 
            0 \\ 0 \\ \sum\limits_{i=1}^{4} \ithfv{i} 
        \end{bmatrix},
    \label{eq:linear_accel_air}\\
    %
    %
    \dot{\rv} &= 
    \begin{bmatrix}
        \cos\theta & 0 & -\cos\phi \sin\theta\\
        0 & 1 & \sin\phi\\
        \sin\theta & 0 & \cos\phi \cos\theta\\
    \end{bmatrix} \qv, 
    \label{eq:angular_speed_air} \\
    \totalInertia\,\dot{\qv}
    &=
    - \dragR \, |\qv| \qv
    - \qv \! \times \! \inertia \qv
    + \wingSpan
        \begin{bmatrix} 
            \ithfv{2}\!-\!\ithfv{4} \\ 
            \ithfv{3}\!-\!\ithfv{1} \\
            -\sum\limits_{i=1}^{4} (-1)^{i} \ithfv{i} 
        \end{bmatrix},
    \label{eq:angular_accel_air}
\end{align}
\noindent where $\Rbtow \in \SO{3}$ denotes Euler rotation matrix. Also, $\totalMass = \mass + \addedMass$ represents the sum of the robot's mass and the added mass effect, respectively. Additionally $\grav$ the gravity vector. The sum of the vehicle's tensor of inertia and the added inertia effect are given by $\totalInertia = \inertia + \addedInertia$. In the air, $\totalMass \approx \mass$ and $\totalInertia \approx \inertia$. Considering the origin of the inertial system and vehicle coincident, and the symmetrical vehicle, the tensor of inertia matrix can be simplified by nullifying the terms outside the main diagonal.

Matrices $\dragP$ and $\dragR$ represent the drag coefficients for translational and rotational motion represented by 
\begin{equation*}
    \dragP = \frac{\density}{2}
    \begin{bmatrix}
        \nabla^{2/3}C_{u}  & 0       & 0       \\
        0       & \nabla 
        ^{2/3}C_{v}  & 0       \\
        0       & 0       & \nabla ^{2/3}C_{w} 
        \label{eq:coef_drag_T}
    \end{bmatrix}, \text{~~and}
\end{equation*}
\begin{equation*}
    \dragR = \frac{\density}{2}
    \begin{bmatrix}
        \nabla^{5/3}  C_{p}  & 0       & 0         \\
        0       & \nabla^{5/3} C_{q}  & 0         \\
        0       & 0       & \nabla^{5/3} C_{r}  
        \label{eq:coef_drag_R}
    \end{bmatrix},
\end{equation*}
\noindent respectively \cite{ConteSerrani1996}. The drag effect depends on the vehicle's attitude since $\vv$ is in $\worldFrame$.

Finally, $\ithfv{i} = \density\,\motorGain \ithOv{i}{}^2$, $\forall i = 1\ldots4$, are the forces generated by the vehicle's motors running with $\ithOv{i}$ speed (in rpm), and rotation moments are dependent of the distance between the rotor and the vehicle's center of mass $\wingSpan$. These forces are proportional to the \emph{thrust coefficient} $\motorGain$, which is a function of geometric parameters of the propeller and physical characteristics of the medium. It is important to observe that both, forces and moments, are dependent on environment density $\density$.

 
To stabilize the vehicle in the aerial operation mode, we used nonlinear or linearized control laws at the operating point $\{\phi,\theta\} \approx 0$ \cite{Drews2014Hybrid}. In such a case, we can define the speed of aerial propellers as:
\begin{equation}
    \Ov = 
    \begin{bmatrix}
        1 & 0 & -1 & 1\\
        1 & 1 & 0 & -1\\
        1 & 0 & 1 & 1\\
        1 & -1 & 0 & -1
    \end{bmatrix} 
    \uv.
\end{equation}

As previously said, $\uv$ represents the unbalancing of forces and moments of equations \eqref{eq:linear_accel_air} and \eqref{eq:angular_accel_air}, respectively, therefore,
\begin{equation}
    \dwT = \frac{\Kp \erro{z} + \Kd \dot{\erro{z}} + \Ki \int \erro{z}}{\cos\phi \cos\theta} + \left(\frac{\mass\|\grav\|}{4 \density \motorGain}\right)^\frac{1}{2},
\end{equation}
\noindent is the nonlinear control law used for stabilizing the robot along the $z$-xis, where the error $\erro{z} = \reference{z}-z$, $\Kp$, $\Ki$ and $\Kd$ are \ac{PID} control gains, and $\density \approx 1$ is the air density.
Also, we can define:
\begin{eqnarray}
    \begin{bmatrix}
        \dwphi\\
        \dwthe\\
        \dwpsi
    \end{bmatrix} = \Kp
    \begin{bmatrix}
        \erro{\phi}\\
        \erro{\theta}\\
        \erro{\psi}
    \end{bmatrix} + \Kd
    \begin{bmatrix}
        \dot{\erro{\phi}}\\
        \dot{\erro{\theta}}\\
        \dot{\erro{\psi}}
    \end{bmatrix} + \Ki
    \int
    \begin{bmatrix}
        \erro{\phi}\\
        \erro{\theta}\\
        \erro{\psi}
    \end{bmatrix},
    \label{eq:attitude_control}
\end{eqnarray}
\noindent as \ac{PID} controllers for the robot's attitude, where the error $\erro{\phi} = \reference{\phi}-\phi$, $\erro{\theta} = \reference{\theta}-\theta$ and $\erro{\psi} = \reference{\psi}-\psi$.
To move the robot along $y$ and $x$-axes, we use $\reference{\phi}$ and $\reference{\theta}$, respectively, such that:
\begin{equation}
    \begin{bmatrix}
        \reference{\phi}\\
        \reference{\theta}
    \end{bmatrix} = \Kp \Rbtow
    \begin{bmatrix}
        \erro{x}\\
        \erro{y}
    \end{bmatrix}
    + \Kd \Rbtow \dot{
    \begin{bmatrix}
        \erro{x}\\
        \erro{y}
    \end{bmatrix}},
\end{equation}
\noindent with the errors $\erro{x} = \reference{x}-x$ and $\erro{y} = \reference{y}-y$. In this context, we can operate the \ac{UAV} in closed-loop by setting the reference vector:
{\setlength{\arraycolsep}{3pt}
\begin{equation*}
    \refv =
    \begin{bmatrix}
        \reference{x} & \reference{y} & \reference{z} & \reference{\psi}
    \end{bmatrix}\transpose{}.
\end{equation*}
}
\subsection{Underwater dynamics and control}
\label{subsec:underwater_dinamics}

Next, we describe the mathematical model for the aquatic operation mode. Since the vehicle operates such a typical \ac{ROV} \cite{Gomes05}, we must define a new input vector $\uv \in \Reais{4}$, such that:
{\setlength{\arraycolsep}{3pt}
\begin{equation*}
    \uv =
    \begin{bmatrix}
        \dwT & \dwV & \dwthe & \dwpsi
    \end{bmatrix}\transpose{},
\end{equation*}
}
\noindent where now $\dwV$ represents the forward thrust.

In the water, the complete dynamics of the robot can be written as:
\begin{align}
    \dot{\pv} &= \vv, 
    \label{eq:linear_speed_wat}\\
    \totalMass \dot{\vv} &=
    - \density \Rbtow \dragP |\vv| \vv
    - \qv \! \times \! \mass \vv
    - \gravTerm
    + \Rbtow 
        \begin{bmatrix} 
            \ithfv{2}\!+\!\ithfv{4} \\
            0 \\
            \ithfv{1}\!+\!\ithfv{3}
        \end{bmatrix},
    \label{eq:linear_accel_wat}\\
    %
    %
    \dot{\rv} &= 
    \begin{bmatrix}
        \cos\theta & 0 & -\cos\phi \sin\theta\\
        0 & 1 & \sin\phi\\
        \sin\theta & 0 & \cos\phi \cos\theta\\
    \end{bmatrix} \qv, 
    \label{eq:angular_speed_wat} \\
    \totalInertia\,\dot{\qv}
    &=
    - \density \dragR \, |\qv| \qv
    - \qv \! \times \! \inertia \qv
    - \gravStability
    + \wingSpan
        \begin{bmatrix} 
            0 \\
            \ithfv{3}\!-\!\ithfv{1} \\
            \ithfv{2}\!-\!\ithfv{4}
        \end{bmatrix},
    \label{eq:angular_accel_wat}
\end{align}
\noindent where the water density is $\density \approx \n{e3}$. Also, $\gravTerm$ and $\gravStability$ are terms that incorporate gravity and buoyancy forces to provide underwater stability, in such a way that the resulting gravity force is:
\begin{equation*}
    \gravTerm = \left(\mass\!-\!\density \vol \right)\!\grav, \text{~~such that~~} \density \vol > \mass,
\end{equation*}
\noindent where buoyancy is slightly greater than weight, and
\begin{equation*}
    \gravStability = \distcgtocb (\mass\!+\!\density \vol)\|\grav\|
    \begin{bmatrix} 
        \sin\phi \\
        \sin\theta \\
        0 
    \end{bmatrix},
\end{equation*}
\noindent is the passive stability moment. These two are both dependent upon the density $\density$ of the medium in which the robot with volume $\vol$ navigates. A restoring moment is ensured by correctly positioning the center of mass (where the gravity vector $\grav$ acts) at a distance $\distcgtocb$ of the center of buoyancy.


In multi-rotor systems, the unbalancing forces and moments generated by the vehicle's thrust systems are the sources of movement. This characteristic is the same in different environments. However, the control strategies of aquatic robots such as \ac{ROVs} are better feasible in the water environment. Both are discussed as follows.

In the underwater operation mode, the force provided by the set of propellers is divided into horizontal and vertical components. Therefore, the speed in the motors is now ruled by:
\begin{equation}
    \Ov = 
    \begin{bmatrix}
        1 & 0 & -1 & 0\\
        0 & 1 & 0 & 1\\
        1 & 0 & 1 & 0\\
        0 & 1 & 0 & -1
    \end{bmatrix}
    \uv.
\end{equation}

Concerning actuation forces and moments, we use the following control law to stabilize the depth of the robot:
\begin{equation}
    \dwT = \frac{\Kp \erro{z} + \Kd \dot{\erro{z}} + \Ki \int \erro{z}}{\cos\phi \cos\theta} - \left( \frac{\|\gravTerm\|}{2 \density \motorGain}\right)^\frac{1}{2}.
\end{equation}

Meanwhile, attitude moments can be stabilized by control laws similar to those in Eq.~\eqref{eq:attitude_control},
\begin{equation}
    \begin{bmatrix}
        \dwthe\\
        \dwpsi
    \end{bmatrix} = \Kp
    \begin{bmatrix}
        \erro{\theta}\\
        \erro{\psi}
    \end{bmatrix} + \Kd
    \begin{bmatrix}
        \dot{\erro{\theta}}\\
        \dot{\erro{\psi}}
    \end{bmatrix} + \Ki
    \int
    \begin{bmatrix}
        \erro{\theta}\\
        \erro{\psi}
    \end{bmatrix} dt,
    \label{eq:attitude_control_water}
\end{equation}
\noindent where $\dwphi$ is assumed to be null since roll moment is self-stabilized.
In this configuration, the robot behaves similarly to a nonholonomic airplane, in a way that its position can be stabilized by a simple kinematic nonlinear control law:
\begin{equation}
    \dwV = \Kv \escalar{d} + \Ka \erro{\psi}^2,
\end{equation}
\noindent where $\escalar{d}$ and $\reference{\psi}$ are the Euclidean distance and the orientation angle between $[\reference{x},\reference{y}]$ and $[x,y]$, respectively.
Similarly to the aerial case, we can define:
{\setlength{\arraycolsep}{3pt}
\begin{equation*}
    \refv =
    \begin{bmatrix}
        \reference{x} & \reference{y} & \reference{z} & \reference{\psi}
    \end{bmatrix}\transpose{}.
\end{equation*}
}

\subsection{Hybrid trajectory planning}

As previously discussed, our hybrid vehicle presents two different operating modes: the aerial mode discussed in Sec.~\ref{subsec:aerial_dynamics}, and the aquatic mode in Sec.~\ref{subsec:underwater_dinamics}. Here, it is assumed an abrupt switch when the vehicle transits from air to water or vice-versa, basically cause by the change in the environment density $\density$.

Problem \ref{prob:main} formalizes the objective as finding a feasible and collision-free trajectory for our \ac{HAWV}, leading it from its starting position to some goal location $\rgoal$. Of course, depending on these initial and final points, and on the obstacle geometries, the vehicle may need to execute a transition between both media. 

It is hard to ensure, from a controller perspective, a smooth transition in such a complex hybrid system. Therefore, in this paper, we decide to work with a hybrid trajectory planner that takes into account the operating modes and the transition zone between them, as presented in Fig.~\ref{fig:model}. This planning strategy computes reference commands for the local control laws that impose a vertical motion mode for the vehicle to dive in or take off from the water.

To solve Prob.~\ref{prob:main}, we proposed a set of algorithms based on the so-called \ac{CL-RRT} \cite{Kuwata2009Realtime}. The \ac{CL-RRT} is a variant of the classic \ac{RRT} approach in which we use a tree $\tree$ composed of nodes $\node[k]$ to explore the obstacle-free portion of the environment. Each node has two components, one that represents reference commands to the low-level control laws, and another that uses the system dynamics to validate the first one. This validation process ensures that the reference $\refv$ is the node can be sent to the controllers of the vehicle without leading him to a state collision.

The \ac{CL-RRT} is generally composed of two procedures: one responsible for expanding the tree through the environment, and another responsible for choosing the current best trajectory $\bestpath$ in the tree and send it to be tracked by the vehicle. In Algorithms \ref{alg:expandtree} and \ref{alg:executionloop}, we present our hybrid version to the \ac{CL-RRT} for hybrid aerial-aquatic vehicles.

Alg.~\ref{alg:expandtree} works as the \ac{RRT}, in the sense that it expands an exploring random tree through the collision-free portions of the space. However, differently from the seminal method proposed by \cite{lavalle2001randomized}, here we modify the system propagation dynamics to incorporate the transition between operating modes due to the change of the major environment.
The algorithm input is the tree $\tree$ to be expanded, the current obstacle-free representation of the environment $\Xfree(t)$ mapped by the robot, the goal $\rgoal$, and the current $t$. 
\begin{algorithm}[!htb]
    \caption{Expand the hybrid tree $\tree$}
    \label{alg:expandtree}
    \begin{algorithmic}[1]
        \REQUIRE $\tree$, $\Xfree(t)$, $\rgoal$, $t$
        \STATE sample $\sv$ from $\Reais{\nrv}$ space (out the transition zone) or choose $\rgoal$
        \label{algline:expandtree:sample}
        %
        \FOR{each node $\node \in \tree$, in the sorted order,}
            \STATE calculate $\refv(t+\Dtc)$ by connecting $\refv_{\node}(t)$ and $\sv$
            \label{algline:expandtree:connect_to_sample}
            \IF {$\refv(t+\Dtc)$ and $\refv_{\node}(t)$ have $z > \transLimiar$}
                \STATE use $\refv(t+\Dtc)$ and the aerial model to propagate $\xv(t)$ obtaining $\xv(t+\Dtc)$ 
                \label{algline:expandtree:propagate_air}
            \ELSIF{$\refv(t+\Dtc)$ and $\refv_{\node}(t)$ have $z < -\transLimiar$}
                \STATE use $\refv(t+\Dtc)$ and the underwater model to propagate $\xv(t)$ obtaining $\xv(t+\Dtc)$ 
                \label{algline:expandtree:propagate_water}
            \ENDIF
            \IF{there is a medium transition between $\xv(t)$ and $\xv(t+\Dtc)$}
                \STATE re-propagate $\refv(t+\Dtc)$ and $\xv(t+\Dtc)$ to take off/dive (vertical) mode
                \label{algline:expandtree:getoff_transition_zone}
                \STATE if it fails, \textbf{goto} line \ref{algline:expandtree:connect_to_sample}
                \label{algline:expandtree:failing_trans}
            \ENDIF
            %
            %
            \IF {$\xv(t) \in \Xfree(t)$ $\forall t \in [t, t\!+\!\Dtc]$} 
                \STATE add intermediate nodes $\node[k]$ to $\tree$
                \label{algline:expandtree:safe_nodes}
                \STATE set appropriate cost-to-go of the trajectory at $\node[k]$
            \ELSIF{$\node[k]$ are feasible}
                \STATE add all $\node[k]$ to $\tree$ and mark them \emph{unsafe} and \textbf{break}
                \label{algline:expandtree:unsafe_nodes}
            \ENDIF
        \ENDFOR
  \end{algorithmic}
\end{algorithm}

The method starts at line \ref{algline:expandtree:sample}, where it samples a configuration $\sv$ from the reference command space. Then, for each node in the tree, it tries to connect the reference command in $\node$ to $\sv$ using a linear propagation during $\Dtc$ seconds. 

\begin{figure}
    \centering
    \includegraphics[width=.8\linewidth]{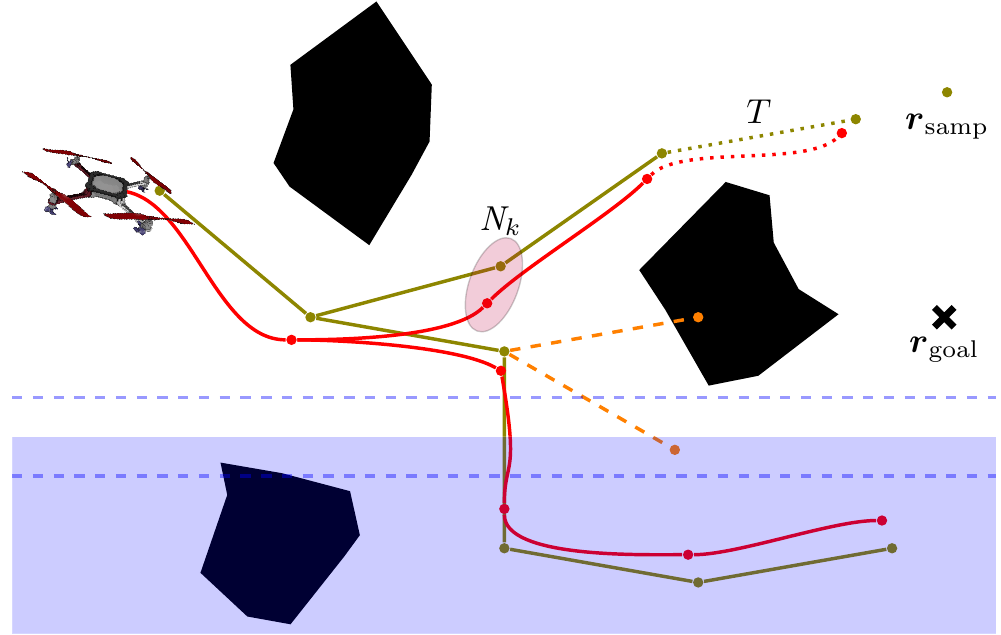}
    \caption{Tree expansion procedure (alg.~\ref{alg:expandtree}): each node $\node$ in the tree is composed of a reference command (yellow branches) and a propagated state (red branches). The tree $\tree$ can only be expanded if, for a given reference, the system realizes a trajectory at $\Xfree$. Infeasible nodes (orange one) are not incorporated into the planner. Also, every node propagated from one medium to another is constrained to a vertical movement that leads the robot out of the transition zone.}
    \label{fig:expansion_tree_example}
\end{figure}

Fig.~\ref{fig:expansion_tree_example} illustrate this expansion process, where the yellow branches are these reference $\refv$ states, while reds branches are the dynamic models propagate using the correspondent yellow ones. 
This propagation happens at lines \ref{algline:expandtree:propagate_air}, \ref{algline:expandtree:propagate_water} or \ref{algline:expandtree:getoff_transition_zone}, depending on the conditions of $\refv(t+\Dtc)$ and $\refv_{\node}(t)$. Here, the main idea is to guarantee that, if the reference command propagated in the tree leads to a media transition, than an special action must be taken. This action is randomly chosen between take off or dive beyond the transition zone, as illustrated in Fig.~\ref{fig:expansion_tree_example}.

The remainder of the algorithm is to validate the new nodes in $\tree$ by testing them against the current map of the space raised by the robot. Nodes that do not lead to colliding with the obstacles are accepted, while others are rejected. Particularly, at line \ref{algline:expandtree:failing_trans}, nodes that fail to perform a vertical movement when passing from one medium to other are also eliminated. By failing, we mean the cases when the vehicle enters or leaves the water surface with $\vv$ along $x$ and $y$ axis significantly different from zero. In order words, when moving in the transition zone, only vertical speed is allowed, and consequently $\reference{\phi} = \reference{\theta} = 0$.

To be able to solve Prob.~\ref{prob:main}, we propose a second algorithm that uses the growing tree to guide the vehicle through the environment in real-time. 
When the mission is performed in dynamic and uncertain environments, the robot must keep growing the exploring tree during the execution cycle to be able to adapt to unpredictable situations \cite{Luders10Bounds}.
Therefore, the procedure called \emph{execution loop} computes the current best trajectory $\bestpath$ at time instant $t$, guiding the robot towards the goal position. Alg.~\ref{alg:executionloop} starts with the initial position of the agent and goes on in the loop that finishes when the main target is achieved.

\begin{algorithm}[!htb]
    \caption{Execution loop}
    \label{alg:executionloop}
    \begin{algorithmic}[1]
        \REQUIRE Initial state $\xv(0)$ and goal reference $\rgoal$
        \STATE initialize $\tree$ with node at $\xv(0)$ 
        \REPEAT
            \STATE update current state $\xv(t)$ and space constraints $\Xfree(t)$
            \label{algline:tree_execution:updatesensors}
            \IF{$z(t) \in \xv(t) > 0$}
                \STATE propagate $\xv(t)$ to $\xv(t+\Dtc)$ by using \eqref{eq:linear_speed_air} to \eqref{eq:angular_accel_air} with aerial controllers
                \label{algline:tree_execution:propagate_x_air}
            \ELSE
                \STATE propagate $\xv(t)$ to $\xv(t+\Dtc)$ by using \eqref{eq:linear_speed_wat} and \eqref{eq:angular_accel_wat}
                \label{algline:tree_execution:propagate_x_water}
            \ENDIF
            \REPEAT
                \STATE expand $\tree$ by using Alg.~\ref{alg:expandtree} 
                \label{algline:tree_execution:expand_tree}
            \UNTIL{time limit $\Dtc$ is reached}
            \STATE choose the current best trajectory $\bestpath \in \tree$ \label{algline:tree_execution:calc_best_path}
            \IF {$\bestpath \in \emptyset$} 
                \STATE apply safety action and \textbf{goto} line \ref{algline:tree_execution:next_time}
            \ENDIF
            \STATE re-propagate from $\xv(t+\Dtc)$ by using references associated with $\bestpath$ 
            \IF {$\xv(t) \in \Xfree(t)$ $\forall t \in [t, t+\Dtc]$} 
                \STATE send best trajectory references to the controller
            \ELSE
                \STATE remove infeasible parts of $\tree$ and \textbf{goto} line \ref{algline:tree_execution:calc_best_path}
            \ENDIF
            \STATE $t \gets t + \Dtc$ \label{algline:tree_execution:next_time}
        \UNTIL{reach $\rgoal$}
  \end{algorithmic}
\end{algorithm}

At line \ref{algline:tree_execution:updatesensors}, the robot uses its sensors to estimate its state and the surrounding environment. It happens every $\Dtc$ period, and it will be used in the three expansion procedure. 
In the sequence, we use the dynamic model of the vehicle to estimate the future state, which will be used to compute the best trajectory while the tree is expanding in line \ref{algline:tree_execution:expand_tree}. However, once we have a hybrid system that behaves differently depending on the environment it is navigating, we must evaluate the altitude value $z(t)$ before compute $\xv(t+\Dtc)$.
When $z(t)$ is higher than zero, that means the vehicle is flying and the aerial model must be used in the prediction (line \ref{algline:tree_execution:propagate_x_water}).

The remainder of the algorithm works like the classic \ac{CL-RRT} \cite{Kuwata2009Realtime}. After expanding the tree, it computes the current best bath and uses it to guide the vehicle in the next iteration of the procedure. Several heuristics can be used to define this trajectory. Here, we use the simple one, that chooses the branch closest to the goal location. 
When such a trajectory does not exist, we generally take a safety action that may be the complete stop of the robot when possible (line \ref{algline:tree_execution:next_time}).  For example, in quadrotor-like vehicles, we can use a hovering mode that allows them to wait until the tree is further extended and a new best trajectory is computed. 

When $\bestpath$ exists, it is possible to use it to re-propagate the current state of the vehicle to evaluate its feasibility. Only then, the references in the best trajectory are sent to the controllers of the real vehicle, and infeasible portions of $\tree$ are removed.

%% file: 04_experiments.tex
\section{Experiments}
\label{sec:experiments}

\begin{figure}[thb]
    \centering
    \includegraphics[width=.5\linewidth]{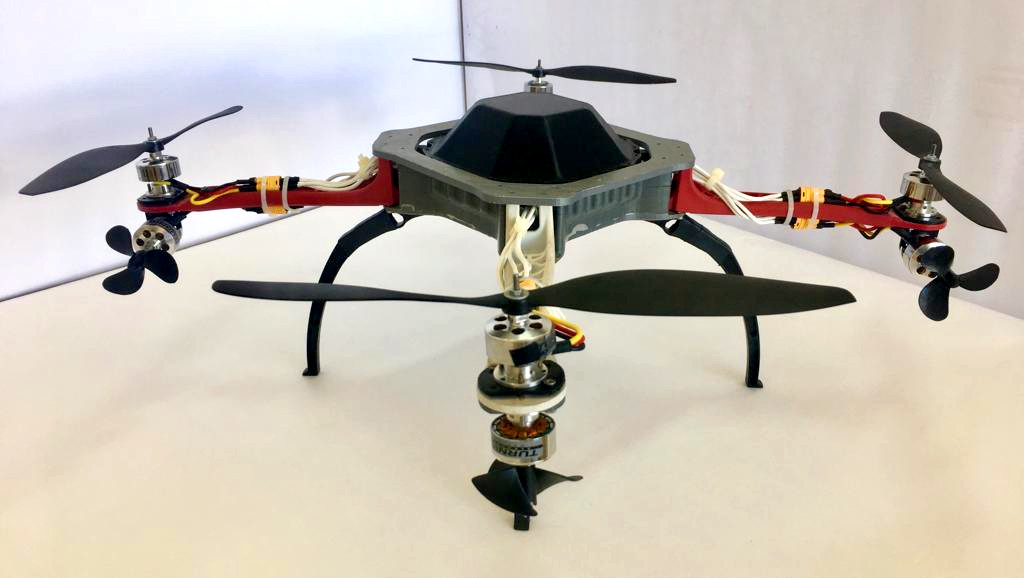}
    \caption{\hydrone~real-world prototype under development.}
    \label{fig:prototype}
\end{figure}

In this section, we present results of our proposed approach. Simulation trials were  executed using Matlab language on an Intel Core$^{\textrm{TM}}$ i7-7500U CPU \n[GHz]{2.70} x 4 and \n[GB]{16} of RAM under Ubuntu 20.04.
The main robot parameters were compiled in Tab.~\ref{tab:parameters}. Most of them were based on a real-world prototype of the \hydrone~(Fig.\ref{fig:prototype}), currently under development on the NAUTEC/FURG, Brazil. The parameters adopted for the aquatic propeller were based on our previous work \cite{Horn2019Study}, while the values for the aerial propeller were extracted from \cite{Maia2017Design}.
{
\nprounddigits{2}
\npfourdigitnosep
\npproductsign{\!\cdot\!}
\renewcommand{\arraystretch}{1.3}
\begin{table}[h]
    \centering
    \caption{Parameters used in the simulated experiments.}
    \label{tab:parameters}
    \setlength{\arraycolsep}{2pt}
    \begin{tabular}{c|cc|cc}
        \hline
        \textbf{Medium} & \textbf{Par.} & \textbf{Values} & \textbf{Par.} & \textbf{Values}\\
        \hline
        \multirow{4}{*}{Both} & \mass~[kg] & \n{1.2860} & \grav~[m/s$^2$] & $\begin{bmatrix} 0 & 0 & \n{9.78} \end{bmatrix}$\\[.2cm]
        & \wingSpan~[m] & \n{0.27} &
        \inertia~[kgm$^2$] & 
        $\begin{bmatrix} 
            \n{1.4465} & \n{.020415} & \n{-.078703}\\
            \n{.020415}  & \n{2.8818} & \n{-.011572}\\
            \n{-.078703} & \n{-.011572} &  \n{1.5410}
        \end{bmatrix} \n{e-2}$\\[.2cm]
        %
        %
        &  \distcgtocb~[m] & \n{.02} & \dragP & $\diagonal{\n{1.25},\n{1.25},\n{4.99}} \n{e-2}$\\[.2cm]
        %
        %
        &  \vol~[m$^3$] & \n{1.6e-3} & \dragR & $\diagonal{\n{1.25},\n{1.25},\n{4.99}} \n{e-2}$\\[.2cm]
        %
        %
        \hline
        \multirow{2}{*}{\shortstack{Air\\$\density \approx \n[kg/m^3]{1.293}$}} &  \totalMass~[kg] & \n{1.2868} & \totalInertia & 
        $\begin{bmatrix}
            \n{1.4466} & \n{.020566} & \n{-.078552}\\
            \n{.020566} & \n{2.8819} & \n{-.011420}\\
            \n{-.078552} & \n{-.011420} & \n{1.5412}
        \end{bmatrix} \n{e-2}$\\[.2cm]
        & \motorGain & \n{2.45e-7} & \motorDrag & \n{5e-11}\\
        %
        %
        \hline
        \multirow{2}{*}{\shortstack{Water\\$\density \approx \n[kg/m^3]{1000}$}} & \totalMass~[kg] & \n{1.9301} & \totalInertia & 
        $\begin{bmatrix}
            \n{1.5639} & \n{0.13781} & \n{.038688}\\
            \n{0.13781} & \n{2.9992} & \n{.10582}\\
            \n{.038688} & \n{.10582} & \n{1.6584}
        \end{bmatrix} \n{e-2}$ \\[.2cm]
        & \motorGain & \n{1.6230e-9} & \motorDrag & \n{1e-11}\\
        \hline
    \end{tabular}
\end{table}
}

All experiments were executed in an environment with dimensions $\n[m]{20} \times \n[m]{20} \times \n[m]{15}$, in which we have randomly distributed 20 spherical obstacles. Since such physical constraints are completely unknown, we endow the robot with the sensory capacity to detect any static obstacle within a radius of up to 3 meters, so that it is able to constantly upgrade $\Xfree(t)$ information.

Finally, the water surface is defined in a way that the water column is \n[m]{5} height. That allows to use $\transLimiar = \n[m]{.8}$, a value at least three times higher than the robot's height.

\begin{figure}
    \centering
    \includegraphics[width=.8\linewidth]{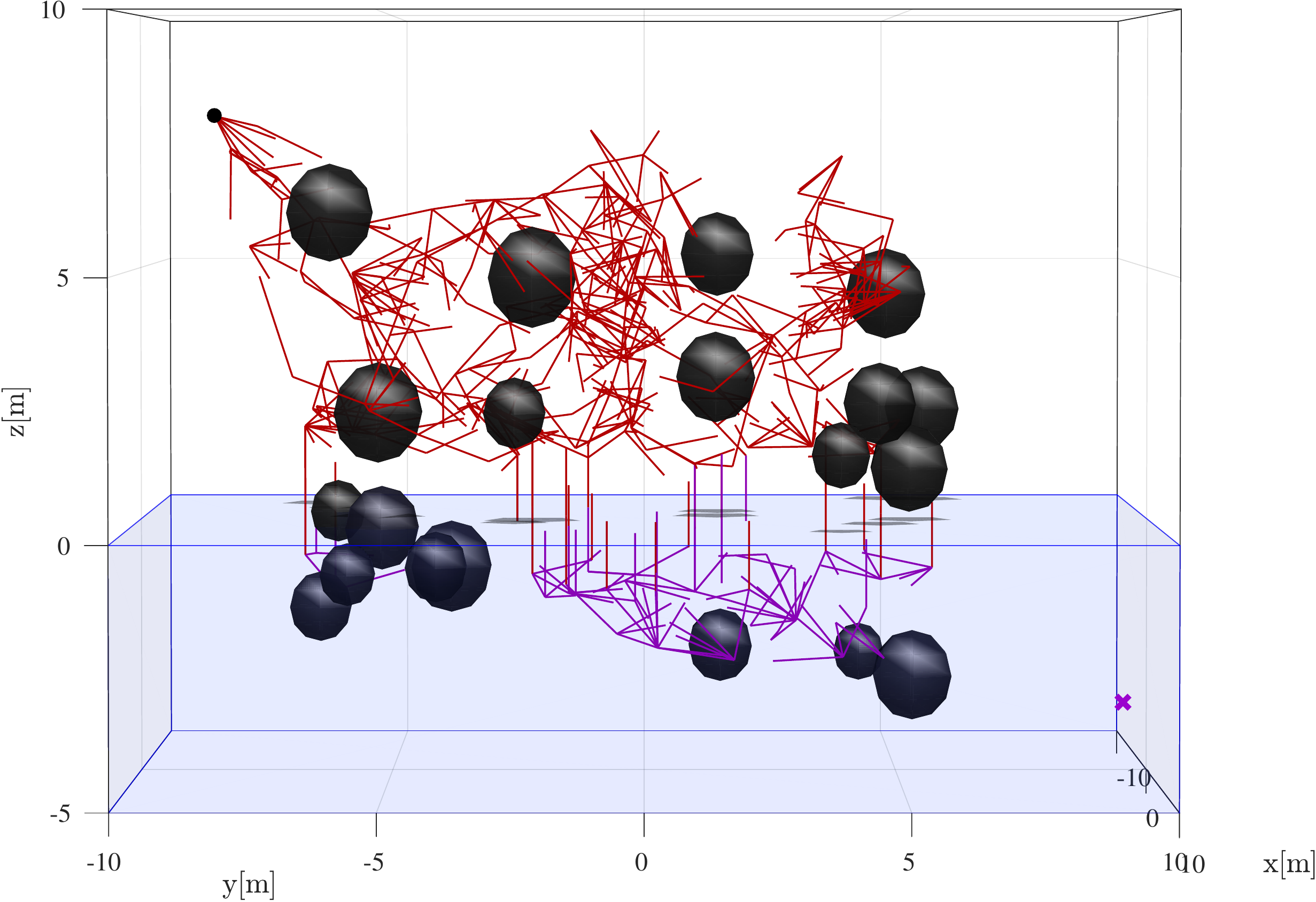}
    \caption{\protect\ac{CL-RRT} planner with transition mode: red nodes represent reference commands valid for the aerial trajectories, while purple nodes are references for the underwater trajectories. The tree root is given by the black dot in the air, while the goal is represented by the purple cross under the water.}
    \label{fig:rrt_transition}
\end{figure}

Fig.\ref{fig:rrt_transition} illustrates an execution of the tree expansion (Alg.~\ref{alg:expandtree}) in the aforementioned environment, with all obstacles known a priory. Here, we only show the reference portions of each node in $\tree$ to demonstrate its hybrid behavior. 
It is possible to notice that all nodes of the tree that promote the transition between water and air make it using a vertical reference propagation. Therefore, we can assure that every best path leading from an aerial location (red branches) to an underwater location (purple branches) is planned by using smooth references.

\subsection{Experiment 1: air to water}

\begin{figure}
    \centering
    \includegraphics[width=.8\linewidth]{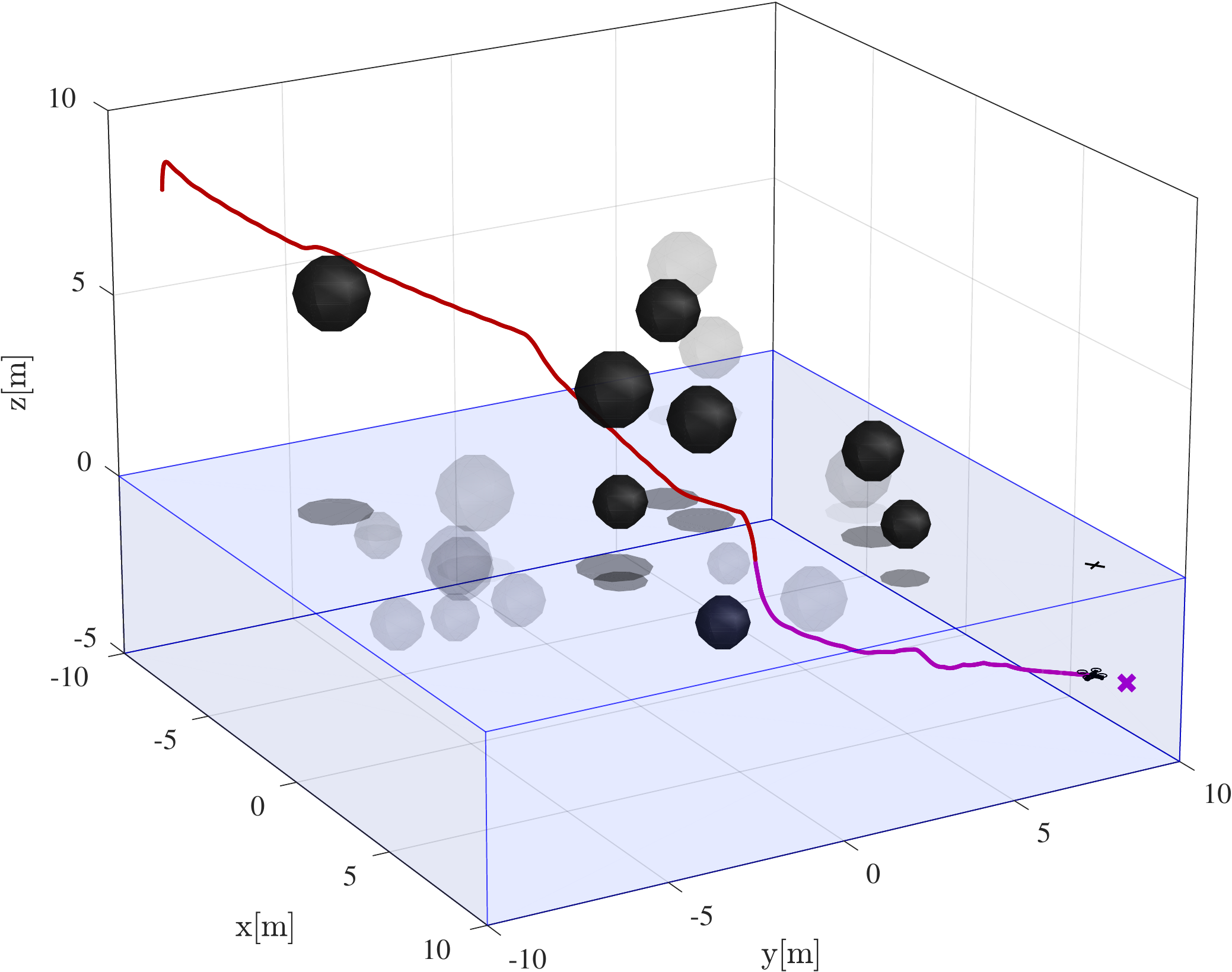}
    \caption{First experiment: navigation in the cluttered environment with transition from air to water. Black spheres represent the obstacles detected by the robot during the loop execution of Alg.~\ref{alg:executionloop}.}
    \label{fig:experiment1}
\end{figure}

In the first experiment, the robot starts at $\pv(0) = \begin{bmatrix} -9 & -9 & ~8 \end{bmatrix}\transpose{}$ with null angles and speeds, and its objective is to reach $\rgoal = \begin{bmatrix} ~9 & ~9 & -3 \end{bmatrix}\transpose{}$. Fig.~\ref{fig:experiment1} presents the complete trajectory executed by the robot until it reaches a little region around the goal. At the beginning, $\Xfree(0) = \Xconj$, since the robot has no prior knowledge about the working space. As it advances in the planned trajectory, more obstacles are detected and incorporates in the planner.

\begin{figure}
    \centering
    \includegraphics[width=.7\linewidth]{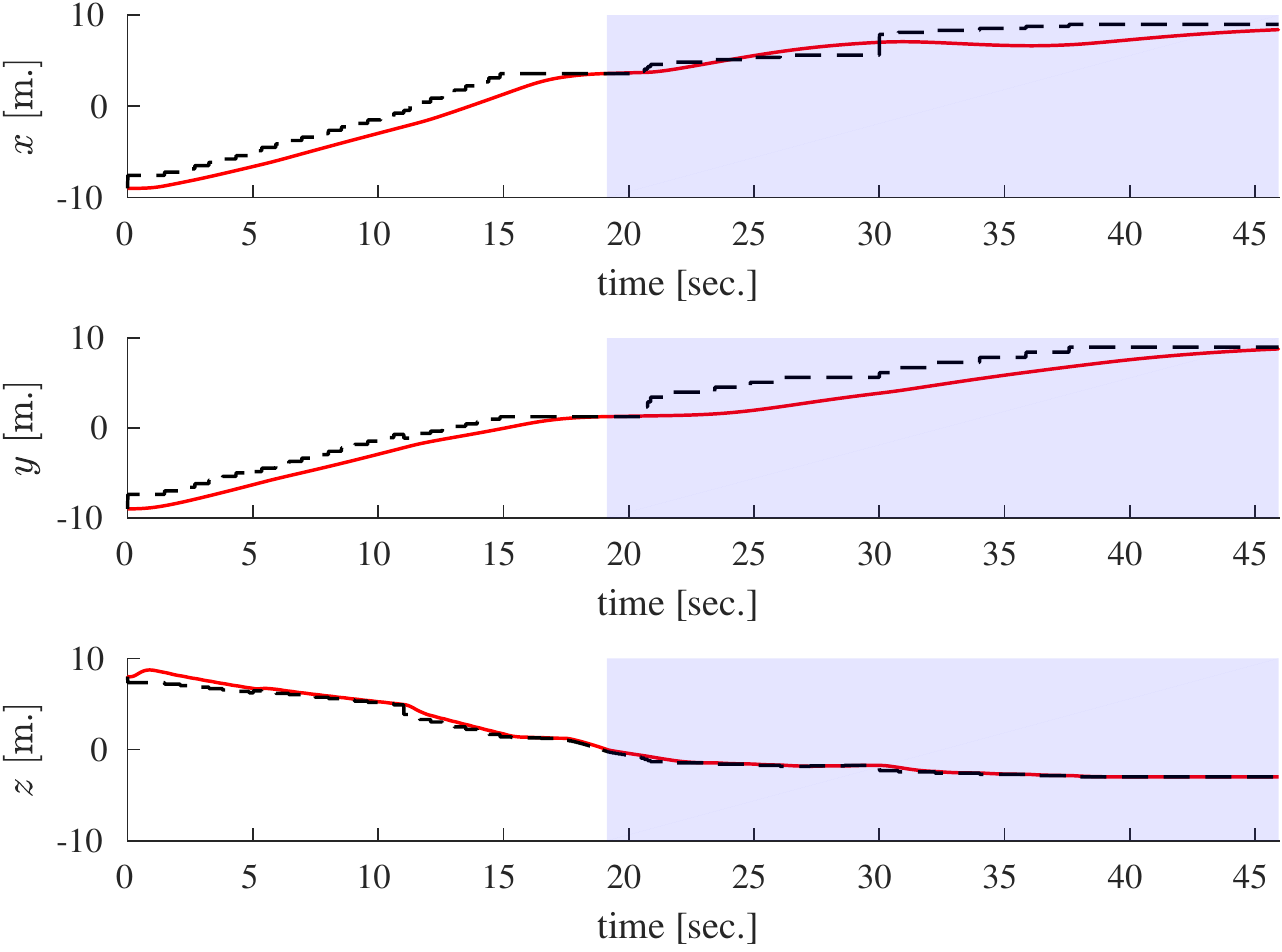}
    \caption{Robot position over time for the first experiment. Red lines are the executed trajectory and the black dot line are the planned one.}
    \label{fig:experiment1_pos}
\end{figure}

\begin{figure}
    \centering
    \includegraphics[width=.7\linewidth]{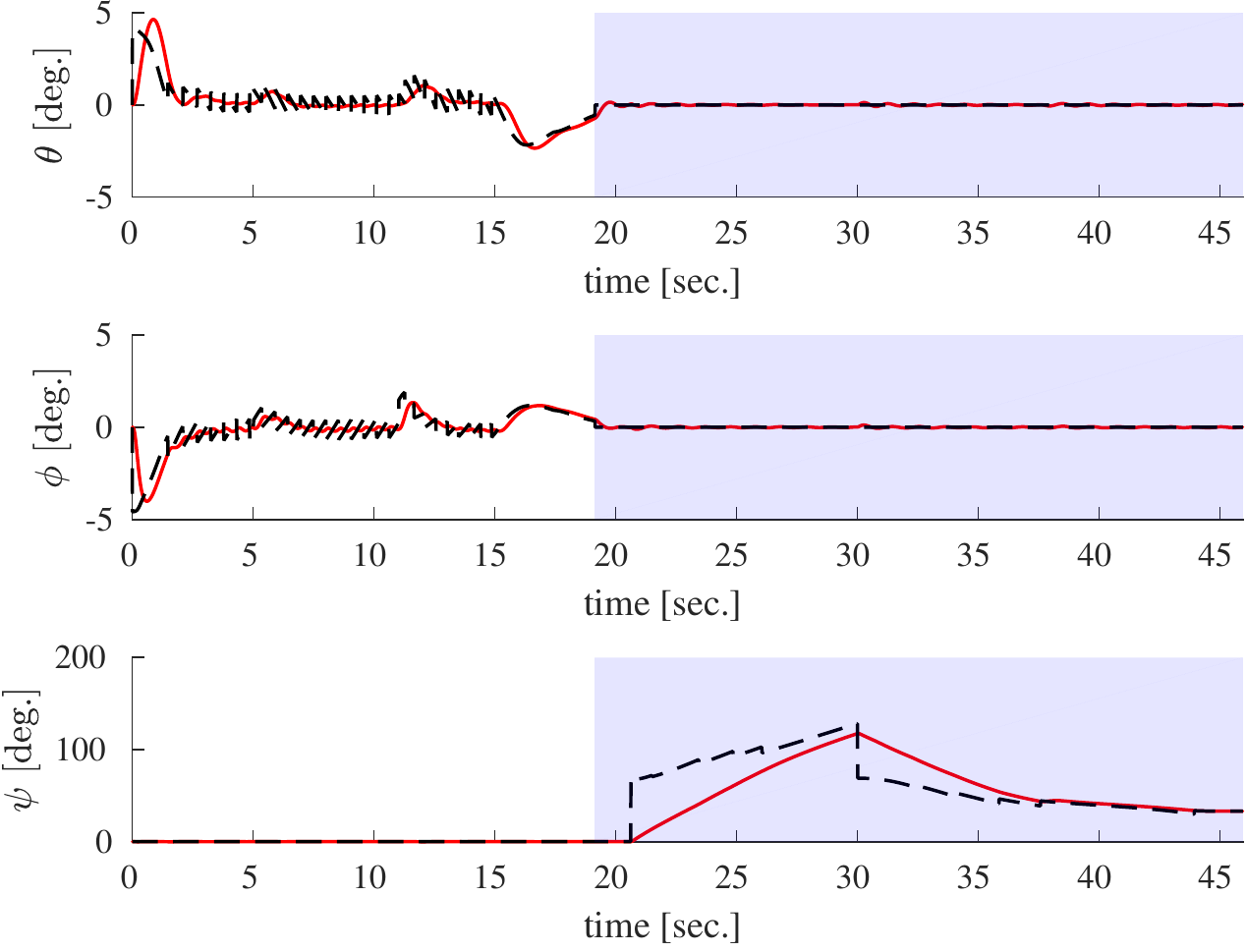}
    \caption{Robot attitude over time for the first experiment. Red lines are the executed trajectory and the black dot line are the planned one.}
    \label{fig:experiment1_ang}
\end{figure}

Figures \ref{fig:experiment1_pos} and \ref{fig:experiment1_ang} present the position and orientation of the vehicle over time for this first experiment. The transition from the air to the water happens at approximately \n[sec]{18} after the start. In that instant, it is possible to perceive that velocities at $x$ and $y$ axes are almost null, while the vehicle descends below $-\transLimiar$. Only when it is outside the transition zone is that the planner is allowed to vary the lateral forces and moments.
This behavior can also be seen in the attitude, once $\phi$ and $\theta$ angles are also near zero when the transition is performed. So, we had a smooth transition media around the operating points of the aerial and aquatic modes.

Also, it is interesting to notice that, although most of the path was traveled by air, the robot spent more time in the water, which can be explained in part by the low speed in the underwater environment. Seeking trajectories that balance flight and submersion time optimizing the robot energy is an interesting challenge for future work.

\subsection{Experiment 2: water to air}

\begin{figure}
    \centering
    \includegraphics[width=.8\linewidth]{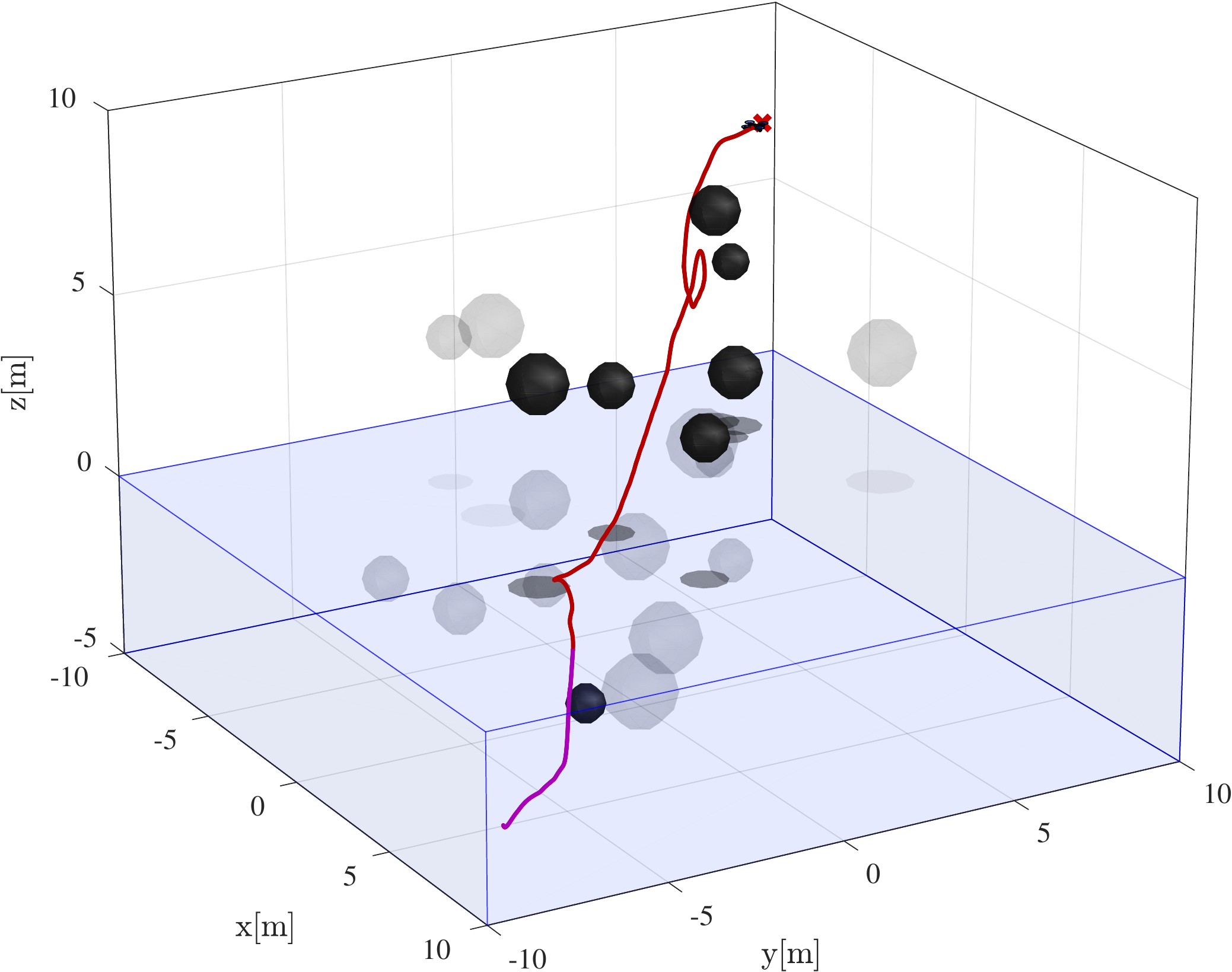}
    \caption{Second experiment: navigation in the cluttered environment with transition from water to air. Black spheres represent the obstacles detected by the robot during the loop execution of Alg.~\ref{alg:executionloop}.}
    \label{fig:experiment2}
\end{figure}

\begin{figure}
    \centering
    \includegraphics[width=.7\linewidth]{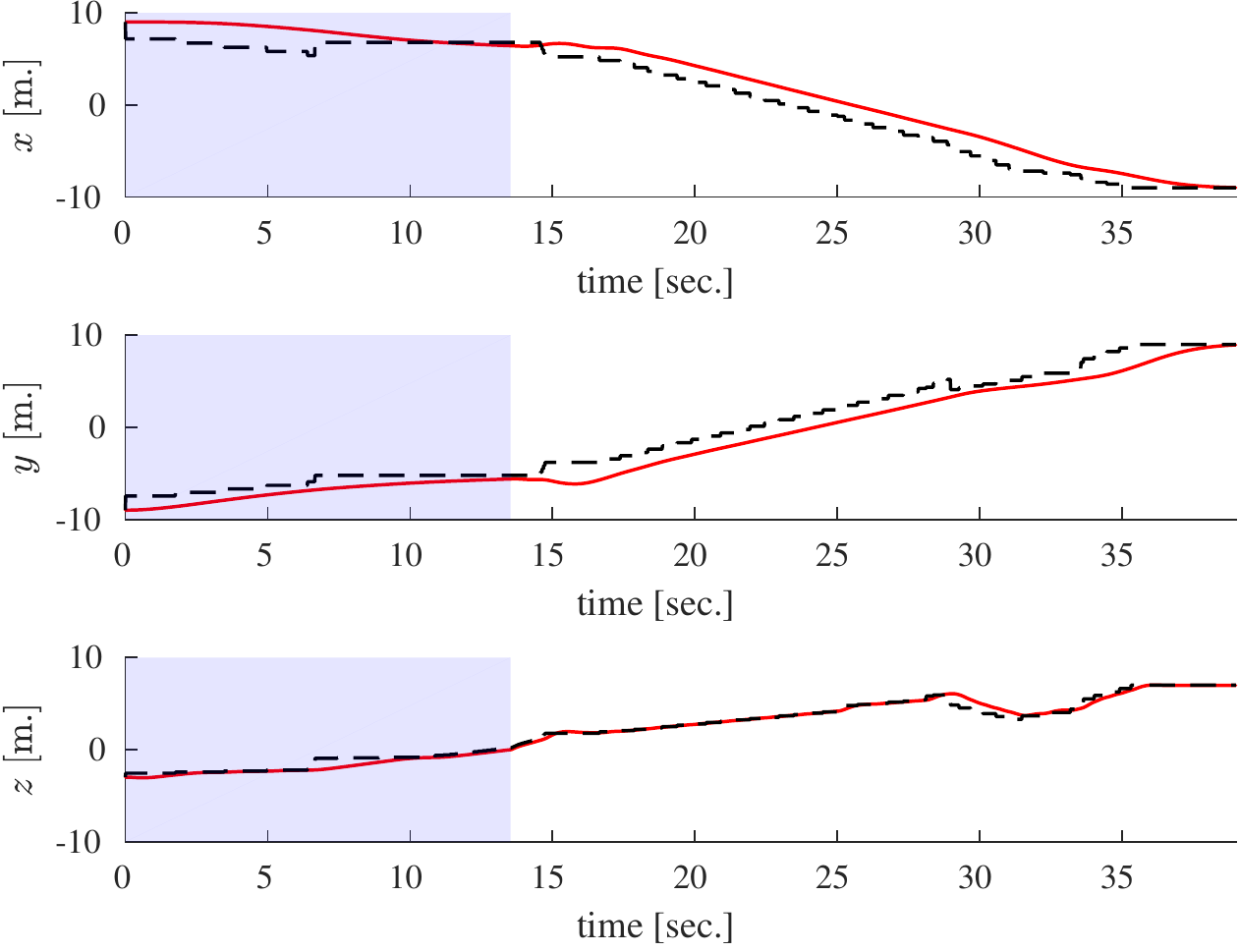}
    \caption{Robot position over time for the second experiment.}
    \label{fig:experiment2_pos}
\end{figure}

\begin{figure}
    \centering
    \includegraphics[width=.7\linewidth]{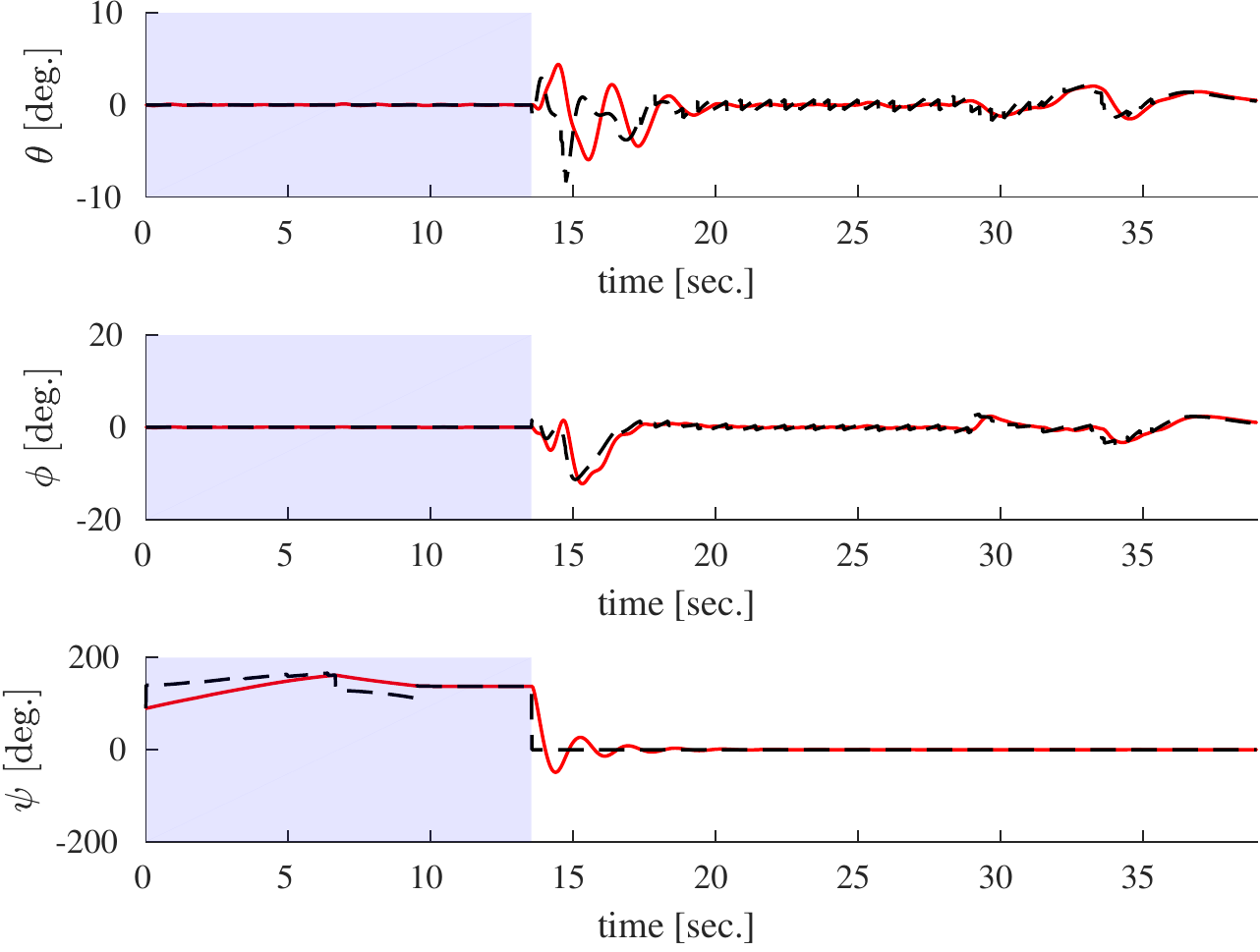}
    \caption{Robot attitude over time for the second experiment.}
    \label{fig:experiment2_ang}
\end{figure}

In the second experiment, the robot has started at $\pv(0) = \begin{bmatrix} ~9 & -9 & -3 \end{bmatrix}\transpose{}$ with null angles and speeds, and its objective is to reach $\rgoal = \begin{bmatrix} -9 & ~9 & ~7 \end{bmatrix}\transpose{}$.
Fig.~\ref{fig:experiment2} presents the complete trajectory, while Figures \ref{fig:experiment2_pos} and \ref{fig:experiment2_ang} show the state behavior of the robot over time. Although the robot has now started underwater, the conclusions of the previous experiment remain valid.
This time, however, the robot remained less time in the water, possibly due to the interference of the first obstacle detected.

%% file: 05_conclusion.tex
\section{Conclusions and future work}
\label{sec:conclusion}

In this paper, we have addressed the problem of real-time trajectory planning for \acd{HAWVs} in unknown cluttered environments. To this end, we have proposed a variant of the \acd{CL-RRT} approach in which we have considered the hybrid dynamics of the system.
Since we have used simple control laws that are unable to ensure stability during the environment change, we have employed a planning step in the tree expansion that forces stable vertical movements to be followed every time the vehicle decides to go from air to water or vice-versa.
Simulated experiments demonstrate the effectiveness of our approach, either from air to water or vice versa. The media transition in all cases was performed in a smooth way, as demonstrated in the states of the vehicle over time. 

In future work, we intend to use control laws more robust to the density switch, in such a way that it will be possible to compute trajectories more aggressive to this kind of hybrid system.
We can also think of some kind of optimization process that will allow the robot not only to compute the trajectory but also decide where is the best point to transit from one medium to another. This evaluation will probably depend on how much energy or time is spent in the air and in the water.

%% file: 06_declarations.tex
\clearpage
\section*{Declarations}

\textbf{Ethical Approval} ~The article has the approval of all the authors.\\

\noindent
\textbf{Consent to Participate} ~All the authors gave their consent to participate in this article.\\

\noindent
\textbf{Consent to Publish} ~The authors gave their authorization for the publishing of this article.\\

\noindent
\textbf{Authors’ contributions:}
\begin{itemize}
    \item \noindent
    \textbf{Pedro Miranda Pinheiro} ~conceived the research, wrote the article, surveyed the literature and contributed to the vehicle modeling and simulation.
    \item \noindent
    \textbf{Armando Alves Neto} ~conceived the research, wrote the article, designed the planning algorithm, collected and processed the test data.
    \item \noindent
    \textbf{Ricardo Bedin Grando} ~wrote the article and surveyed the literature.
    \item \noindent
    \textbf{César Bastos da Silva} ~wrote the article and contributed to the vehicle modeling and simulation.
    \item \noindent
    \textbf{Vivian Misaki Aoki} ~wrote the article and contributed to the vehicle modeling and simulation.
    \item \noindent
    \textbf{Dayana Santos Cardoso} ~wrote the article and contributed to the vehicle modeling and simulation.
    \item \noindent
    \textbf{Alexandre Campos Horn} ~proposed the vehicle and contributed to the modeling.
    \item \noindent
    \textbf{Paulo Lilles Jorge Drews Jr.} ~conceived the research, wrote the article and discussed the main ideas of the article.
\end{itemize}

\noindent
\textbf{Funding} ~National Council for Scientific and Technological Development (CNPq), Coordination for the Improvement of Higher Education Personnel (CAPES) and Human Resources Program from National Agency of Petroleum, Natural Gas and Biofuels (PRH-ANP).\\

\noindent
\textbf{Competing interest} ~There are not conflict of interest or competing interest.\\

\noindent
\textbf{Code availability} ~Not applicable.\\\textbf{}